\documentclass[sigconf,nonacm]{aamas}
\makeatletter
\gdef\acmConference@shortname{\color{white}}
\gdef\acmConference@date{}
\gdef\acmConference@venue{}
\makeatother

\usepackage{balance} 
\newcommand{\tx}{\mathrm}

\title[Learned Terminal Costs for Truck Highway Control]{Learned Terminal Costs for Short-Horizon Model Predictive Control of Autonomous Trucks in Interactive Highway Traffic}

\author{Hooman Hamze}
\authornote{Both authors contributed equally to this research.}
\affiliation{
  \institution{Chalmers University of Technology}
  \city{Gothenburg}
  \country{Sweden}}
\email{hoomanh@chalmers.se}
\email{hooman.hamze230@gmail.com}

\author{Benjamin Houshmand}
\authornotemark[1]
\affiliation{
  \institution{Chalmers University of Technology}
  \city{Gothenburg}
  \country{Sweden}}
\email{benhou@chalmers.se}

\author{Deepthi Pathare}
 \affiliation{
  \institution{Chalmers University of Technology and University of Gothenburg}
  \city{Gothenburg}
  \country{Sweden}}
  \affiliation{
  \institution{Trucks Technology and Industrial Division, Volvo Group}
  \city{Gothenburg}
  \country{Sweden}}
\email{pathare@chalmers.se}

\author{Erik Börve}
\affiliation{
  \institution{Chalmers University of Technology}
  \city{Gothenburg}
  \country{Sweden}}
  \affiliation{
  \institution{Trucks Technology and Industrial Division, Volvo Group}
  \city{Gothenburg}
  \country{Sweden}}
\email{borerik@chalmers.se}

\author{Morteza Haghir Chehreghani}
\affiliation{
  \institution{Chalmers University of Technology and University of Gothenburg}
  \city{Gothenburg}
  \country{Sweden}}
\email{morteza.chehreghani@chalmers.se}

\author{Nikolce Murgovski}
\affiliation{
  \institution{Chalmers University of Technology}
  \city{Gothenburg}
  \country{Sweden}}
\email{nikolce.murgovski@chalmers.se}

\begin{abstract}
Autonomous highway driving for heavy-duty trucks is a challenging control problem. The greater mass, dimensions and intricate dynamics increase the requirements for cooperation from surrounding traffic in critical maneuvers, requiring a careful balance between safety constraints and efficient long-horizon decision making. Model Predictive Control (MPC) provides rigorous constraint satisfaction but relies on terminal cost which can limit its effectiveness at short, computationally affordable horizons. Reinforcement Learning (RL) captures long-term driving behavior but offers no constraint guarantees and tends toward overly conservative policies when trained under sparse collision penalties. In this paper, we propose a hybrid MPC and RL framework for highway driving in autonomous trucks, where a finite-horizon MPC, solved via dynamic programming, uses a pre-trained RL model for terminal cost estimation. We evaluate the framework in a simulation environment with reactive surrounding traffic and compare it against pure MPC with zero terminal cost, rollout with terminal cost estimation using a base policy, and pure RL. Results show that the hybrid approach achieves zero collisions with the lowest cost of any configuration.
\end{abstract}


\newcommand{\BibTeX}{\rm B\kern-.05em{\sc i\kern-.025em b}\kern-.08em\TeX}

\begin{document}




\maketitle 


\section{Introduction}

Autonomous driving promises three concrete benefits over human-driven vehicles: improve safety, since the large majority of road accidents are caused by human error rather than mechanical failure \cite{singh2015critical}; cost efficiency, since speed and fuel consumption profiles can be optimized directly rather than left to individual driving style \cite{s21196547}; and continuous utilization, since an automated vehicle is not bound by mandatory rest periods, fatigue, or shift limits and can in principle operate around the clock \cite{turnbull2017hours}. These benefits are amplified for heavy-duty commercial vehicles such as long-haul trucks, where driver wages are the largest single component of operating cost \cite{leslie2025analysis} and where even small efficiency gains scale directly with fleet size.

Achieving a reliable autonomous vehicle is, however, considerably harder for heavy vehicles than for passenger cars. A truck pulling a trailer has coupled dynamics that a single rigid-body vehicle does not. The trailer does not track the tractor's path directly, which makes both motion planning and the underlying control problem more involved \cite{Borve2023}. Its size and limited maneuverability mean that routine highway actions such as lane changes and merges are harder and require more cooperation from surrounding traffic \cite{9750012}. Highway driving is, in addition, inherently interactive for any vehicle. An agent's own actions influence what nearby drivers do next, so a controller that reacts only to the present state, without any model of how the situation might unfold, tends to be either too timid to be useful or too naive to be safe \cite{9750012}. Meeting all of these requirements at once is the central difficulty this paper addresses.

Two methodological families dominate the literature on autonomous vehicle decision-making and control: Model Predictive Control (MPC) and Reinforcement Learning (RL) \cite{REITER2026101045}. MPC formulates driving as a constrained optimization problem. Given a model of the vehicle and its surroundings, solve for the control sequence that minimizes a cost over a finite look-ahead horizon subject to constraints on speed, acceleration, steering, and collision avoidance, then apply only the first action and re-solve at the next step. Beyond the visible horizon, the controller must fall back on an approximate terminal cost standing in for everything that happens afterwards. A short horizon with a crude terminal cost is fast to solve but can behave poorly, while a longer horizon is more reliable but increasingly expensive online. RL, by contrast, learns a decision policy directly from interaction, with no requirement that the dynamics be known, which suits the strategic aspects of driving. However, a learned policy offers no guarantee its output will respect a physical or safety constraint, and its behavior outside the situations it was trained on is unpredictable, which has motivated explicit
out-of-distribution detection for learned driving
policies~\cite{Hoel2020}.

Several works rely on MPC alone for interactive highway maneuvers. Börve et al. \cite{Borve2023} couple an autonomous truck's trajectory optimization with an iterative update against a traffic predictor to capture genuine interactivity during forced lane changes, but at the cost of solving several parallel nonlinear optimizations every control step, with no formal convergence guarantee and a terminal cost fixed by design. Brito et al. \cite{9750012} use MPC with a hand-tuned, fixed velocity reference for merging in dense traffic and report a markedly higher collision rate than the same architecture using a learned reference instead.

RL alone has also been used directly for tactical driving decisions. Acquarone et al. \cite{Acquarone2023} apply a TD3 agent to cooperative adaptive cruise control for heavy-duty battery electric vehicles, achieving substantial energy savings by exploiting air-drag reduction in a platoon, though the controller is longitudinal only and does not address lane-level decisions. Hoel et al. \cite{Hoel2018} trained a Double-DQN agent to choose discrete lane-change and speed actions for a truck-trailer combination, but like every pure-RL controller, without formal safety guarantees at all. Later in \cite{Hoel2020}, they proposed a bayesian RL technique that estimates uncertainty of decisions to choose safe actions, measurably reducing collisions outside the training distribution at the cost of a hand-set threshold and roughly double the training time. Pathare et al. \cite{Pathare2023} separated high level and low decision making between RL and physics based controllers, which improved safety and driving efficiency relative to letting RL act directly. However, the fixed rule-based execution layer carries none of MPC's constraint guarantees.

Motivated by these complementary shortcomings, a number of recent works combine MPC and RL. A first group targets MPC's tunable parameters. Gupta et al.~\cite{Gupta2024} add an RL-trained throttle correction in parallel with an MPC. Chen et al.~\cite{Chen2024} use RL to choose the prediction horizon for single-vehicle path tracking beating fixed-horizon baselines, but without any surrounding traffic. Zarrouki et al. \cite{Zarrouki2024} pre-compute a catalog of cost-function weight sets that are each individually verified safe, and let an RL agent only ever choose among them for vehicle motion control. Wang et al.~\cite{Wang2023} let RL tune a single MPC weight conditioned on a risk classification for car-following. A second group interfaces a learned component with MPC at the level of a discrete decision or reference trajectory. Mallick et al.~\cite{Mallick2025} propose a learned gear-shift schedule fed into a continuous MPC. Reiter et al. \cite{Reiter2022} let an RL policy reshape an MPC's entire reference trajectory and cost weighting for autonomous racing, beating both a non-strategic MPC and an end-to-end RL policy while remaining safe. Albarella et al.~\cite{albarella2023hybrid} combine a Double-DQN decision layer with two NMPCs for highway driving, though the NMPC terminal cost stays fixed. In all of these, learning never touches the terminal cost itself.

A separate line of work uses RL to approximate the terminal cost directly. Bertsekas~\cite{Bertsekas2024} provides the theoretical foundation, showing that one step of MPC look-ahead on top of an approximate terminal cost is equivalent to one Newton step on the Bellman equation, so even a mediocre approximation yields a near-optimal policy after sufficient look-ahead. Building on this, Morita et al.~\cite{Morita2024} use a learned value function as the terminal cost of an MPPI controller for a robot; Lin et al.~\cite{Lin2024} and Moreno-Mora et al.~\cite{MORENOMORA20233874} learn it via policy and value iteration respectively for general nonlinear systems; and Kang et al.~\cite{Kang2022} learn it online for a robotic manipulator. All of these are validated on continuous low-level control in single-agent, non-reactive settings, and none addresses the discrete tactical decisions or interactive multi-agent traffic that motivate the present work.

Two limitations cut across this literature. First, every learned-terminal-cost method operates on continuous low-level control; none addresses a discrete tactical decision such as a lane change, where the terminal cost must interact with a combinatorial DP optimization. Second, the entire terminal-cost-learning line is validated only in single-agent, non-reactive settings. It is not tested in an interactive environment where surrounding vehicles respond to the ego and thereby shift the very state distribution the value function was trained on. Conversely, the hybrid schemes that do operate in traffic~\cite{Chen2024, Zarrouki2024, Wang2023, Gupta2024, Mallick2025} confine learning to horizon selection, a handful of cost weights, or an additive correction, and never learn the terminal cost that governs beyond-horizon foresight. No prior work integrates a learned or even rollout-based terminal cost into a MPC for a discrete tactical highway-driving problem, nor evaluates whether such a terminal cost remains effective when surrounding traffic is genuinely reactive.


We develop a highway controller for an autonomous truck that solves its finite-horizon MPC problem by dynamic programming over a discretized state-action space, and replaces the usual hand-designed terminal cost with a DQN-learned terminal cost, trained offline and inserted directly into the DP backward pass. For the purpose of benchmarking, a rollout terminal cost approximations, obtained by simulating a working base policy forward from the end of the horizon is also developed. Both retain MPC's exact constraint handling and receding-horizon replanning. Our contributions are as follows:
 
\begin{enumerate}
    \item We propose a hybrid MPC and RL framework for autonomous highway driving of a heavy-duty truck. The MPC performs speed control and lane-change decisions over a discrete state-action grid, while a pre-trained Deep Q-Network (DQN) provides the terminal cost to improve long-term decision making.
 
    \item We evaluate the effectiveness of the proposed framework in a highway traffic simulation by comparing it with conventional MPC and rollout-based terminal cost methods. The proposed DQN-based terminal cost consistently improves safety and decision consistency under both deterministic and noisy observation settings.
 
    \item We demonstrate that incorporating a learned terminal cost enables short-horizon MPC to achieve the best balance between safety, driving performance, and computational efficiency. The proposed three-step MPC eliminates the need for computationally expensive long-horizon optimization while remaining suitable for real-time implementation.
\end{enumerate}

\section{PROBLEM FORMULATION}
\label{Preliminaries}

We consider an autonomous heavy truck operating on a multi-lane highway in the presence of surrounding traffic. Surrounding vehicles are not controlled by the ego truck, but respond to its actions through their own car-following and lane-changing models~\cite{krauss1998microscopic, erdmann2015lane}, so the environment is interactive and the optimal policy depends on multi-step predictions of how traffic will evolve.

\subsection{System Model}
 
The ego truck's state at step $k$ is:
\begin{equation}
    x^e_k = [s^e_k,\, l^e_k,\, v^e_k,\, c^e_k]^\top
\end{equation}
where $s^e_k$ is longitudinal position, $l^e_k \in \{0,\ldots,N_\tx{lanes}-1\}$ is lane index and $N_\tx{lanes}$ is the number of lanes, $v^e_k \in [v^e_\tx{min}, v^e_\tx{max}]$ is velocity with $v_{\text{min}}^{e}$ and $v_{\text{max}}^{e}$ as its bounds, and $c^e_k$ is a lane-change counter that enforces transition and cooldown durations. The control input:
\begin{equation}
    u^e_k = [a^e_k,\, \Delta l^e_k]^\top
\end{equation}
pairs a discretized longitudinal acceleration $a^e_k$ with a discrete lane-change command $\Delta l^e_k \in \{-1, 0, 1\}$. The ego dynamics
\begin{equation}
    x^e_{k+1} = f(x^e_k, u^e_k)
\end{equation}
evolve under a discrete-time kinematic model; the lane and counter update jointly, so a new lane-change command is accepted only once the counter has cleared its transition time and the ego truck reaches its target lane. Section~\ref{Cost Function} provides more details on the lane-change command and the counter.
 
Surrounding vehicles have states $x^T_k = \{x^T_{j,k}\}_{j \in \mathcal{N}_k}$, where $\mathcal{N}_k$ is the set of vehicles within sensor range at step $k$. Their future states $x^T_i$ at step $i > k$ are uncertain for two reasons. First, the initial observations $x^T_k$ are subject to sensor measurement noise. Second, even with perfect initial observations, surrounding vehicles' future behavior can only be approximated by a simplified prediction model. They respond to each other and to the ego in ways that are not fully predictable. Both sources of uncertainty accumulate over the prediction horizon and must be accounted for in the controller design. A detailed description of the prediction model is provided in Section~\ref{Trajectory Prediction}.

\subsection{The MPC Problem}
\label{Model Predictive Control}

The fundamental control objective is to find a policy $\pi$ minimizing the expected cumulative cost over the full traffic state $x_k = (x^e_k, x^T_k)$:

\begin{equation}
\label{eq:ideal}
   \inf_{\pi}\; \mathbb{E}\!\left[\sum_{k=0}^{\infty} \ell(x_k, u_k)\right]
\end{equation}
where the expectation is over the stochastic evolution of surrounding traffic. Solving Eq.~\eqref{eq:ideal} directly is intractable because it requires optimizing over an infinite sequence of future actions under physical constraints and uncertain traffic. MPC provides a tractable approximation by truncating the optimization horizon to $N$ steps and re-optimizing at every step using the latest observations. Given the discretized state and action spaces, the resulting finite-horizon problem can be solved online using Dynamic Programming (DP). However, performing the DP backward pass over the full joint state space $(x^e_k,x^T_k)$ would be intractable, as the state grid grows exponentially with the number of surrounding vehicles~\cite{bertsekas2012dp2}. The DP is therefore performed over the ego vehicle's state space $x^e$, while the predicted surrounding vehicle states are treated as time varying parameters during the optimization. This is reflected in the conditional notation $\ell(x^e_i, u^e_i \mid x^T_i)$: the ego trajectory is optimized while the surrounding states enter as given parameters at each step.

Because the surrounding vehicle states $x^T_i$ are uncertain, we formulate the control task as a chance-constrained MPC~\cite{7740982}. At each step $k$, the controller solves:

\begin{subequations}
\label{eq:optimization}
\begin{align}
    &\min_{u^e_0,\ldots,u^e_{N-1}} \quad
      \mathbb{E}\!\left[\sum_{i=0}^{N-1} \gamma^i\,\ell(x^e_i, u^e_i \mid x^T_i)
      + \gamma^N V_f(x^e_N \mid x^T_N)\right] \\
    &\text{s.t.} \quad x^e_{i+1} = f(x^e_i, u^e_i), \\
    &\phantom{\text{s.t.}} \quad
      \mathbb{P}\!\left[x^e_i \in \mathbb{X}^T(x^T_i)\right]
      \geq 1 - \epsilon, \label{eq:col_avoid} \\
    &\phantom{\text{s.t.}} \quad
      x^e_i \in \mathbb{X}^e, \quad u^e_i \in \mathbb{U}^e, \\
    &\phantom{\text{s.t.}} \quad
      x^e_0 = x^e_k, \quad
      \forall\, i \in \{0,\ldots,N-1\}
\end{align}
\end{subequations}

where $\ell(x^e_i, u^e_i \mid x^T_i) \geq 0$ is a stage cost penalizing undesired ego actions and states, conditioned on surrounding vehicle states; $V_f(\cdot)$ approximates the cost-to-go beyond the horizon; $\mathbb{X}^T(x^T_i)$ is the set of ego states that are safe with respect to the surrounding vehicles at step $i$; $\mathbb{X}^e$ and $\mathbb{U}^e$ encode state and action limits; $\epsilon \in (0,1)$ is the risk tolerance for collision avoidance; and $\gamma \in (0,1]$ is a discount factor. Unlike the undiscounted objective common in classical MPC, $\gamma^i$ down-weights stage costs at later horizon steps. Section~\ref{Discounted Objective} discusses the motivation for this choice.

Only the first optimal action $u^{e*}_0$ is applied, and the state is then re-measured and the problem re-solved at $k+1$. This corrects automatically for prediction errors as new observations arrive.

The chance constraint Eq.~\eqref{eq:col_avoid} treats $x^T_i$ as a random variable incorporating both measurement noise and prediction model uncertainty, and requires the ego to remain safe with probability at least $1-\epsilon$. Under a Gaussian uncertainty model, this constraint admits a deterministic equivalent~\cite{7740982} in which the required safe distance is inflated by a margin proportional to the predicted position standard deviation. Suppressing that margin recovers a deterministic controller acting on the nominal prediction $\bar{x}^T_i$, which serves as a baseline to isolate the benefit of uncertainty handling.

The design burden of Eq.~\eqref{eq:optimization} is the terminal cost $V_f$. Classical stability theory~\cite{MAYNE2000789} requires $V_f$, an associated terminal constraint set $\mathbb{X}^e_f$, and a local controller to jointly satisfy a set of Lyapunov-type conditions, which is already difficult to construct for general nonlinear systems~\cite{rawlings2020model}. In the present setting, this difficulty is compounded by the presence of surrounding vehicles. The terminal set $\mathbb{X}^e_f$ must be control-invariant under the terminal controller, but invariance cannot be guaranteed when surrounding vehicles are free to arbitrarily enter or leave the ego's vicinity beyond the planning horizon.

Solving Eq.~\eqref{eq:optimization} by DP makes it straightforward to substitute alternative terminal cost approximations $V_f(x^e_N\mid x^T_N)$ into the same backward pass. 

\subsection{The Terminal Cost Approximation Problem}
\label{Reinforcement Learning and the Bellman Equation}

The quantity that $V_f$ approximates can be characterized exactly. Viewing the same task as a Markov Decision Process (MDP)~\cite{sutton1998reinforcement}, the agent at state $x^e_k$ selects $u^e_k$ under a policy $\pi$, incurs cost $\ell(x^e_k, u^e_k \mid x^T_k)$, and transitions to $x^e_{k+1}$. The value function of $\pi$,
\begin{equation}
V^{\pi}(x^e) = \mathbb{E}\!\left[\sum_{k=0}^{\infty}\gamma^k\,
\ell(x^e_k, u^e_k \mid x^T_k) \;\middle|\; x^e_0=x^e,\; u^e_k=\pi(x^e_k)\right]
\end{equation}
satisfies the Bellman equation:
\begin{equation}
    V^{\pi}(x^e) = \ell(x^e,\pi(x^e) \mid x^T) + \mathbb{E}\!\left[\gamma\,V^{\pi}(f(x^e,\pi(x^e)) \mid x^T)\right]
\end{equation}
and the optimal value function minimizes over the action at every state:
\begin{equation}
\label{eq:bellman_opt}
V^{*}(x^e) = \min_{u^e}\bigl\{\ell(x^e,u^e \mid x^T)
+ \mathbb{E}\!\left[\gamma\,V^{*}(f(x^e,u^e) \mid x^T)\right]\bigr\}
\end{equation}

Setting $V_f = V^{*}$ in Eq.~\eqref{eq:optimization} makes the truncated and unconstrained (we cannot exactly compute the target cost if constraints are imposed beyond the horizon) $N$-step problem exact, and its solution coincides with the optimal infinite-horizon policy for any $N \geq 1$, since $V^{*}$ accounts precisely for all cost incurred beyond the horizon. This holds because Eq.~\eqref{eq:optimization} is the Bellman recursion unrolled $N$ times, which requires the discount factor of the MPC objective, the discounting applied to the terminal cost, and the discount factor defining $V^{*}$ to be identical; a mismatch would make the truncated problem a different optimization rather than an exact reformulation. $V^{*}$ is therefore the ideal terminal cost, and it encodes exactly the information the MPC lacks beyond step $N$.

However, for the constrained system considered here, $V^{*}$ cannot be obtained. The dynamics depend on surrounding vehicles whose behavior is not modeled in closed form, and the recursion Eq.~\eqref{eq:bellman_opt} admits no analytical solution. The terminal cost must therefore be approximated. Hence, the design question becomes how much approximation error can be tolerated. Bertsekas~\cite{Bertsekas2024} shows that solving Eq.~\eqref{eq:optimization} with $V_f \approx V^{*}$ and applying only the first action is equivalent to one Newton step on the Bellman optimality equation Eq.~\eqref{eq:bellman_opt}. The closed-loop policy is consequently more accurate than $V_f$ itself since an imperfect terminal cost is substantially corrected by even a few steps of look-ahead, and the residual error decreases sharply as $N$ increases. A short horizon is thus sufficient provided $V_f$ is a reasonable approximation of $V^{*}$, which is what makes the terminal cost, rather than the horizon length, the effective design variable.

Section ~\ref{Proposed Method} presents an RL-based model to obtain $V_f \approx V^*$, to be used directly in the DP backward pass as the terminal cost.


\section{Proposed Method}
\label{Proposed Method}

\begin{figure}
    \centering
    \includegraphics[width=1\linewidth]{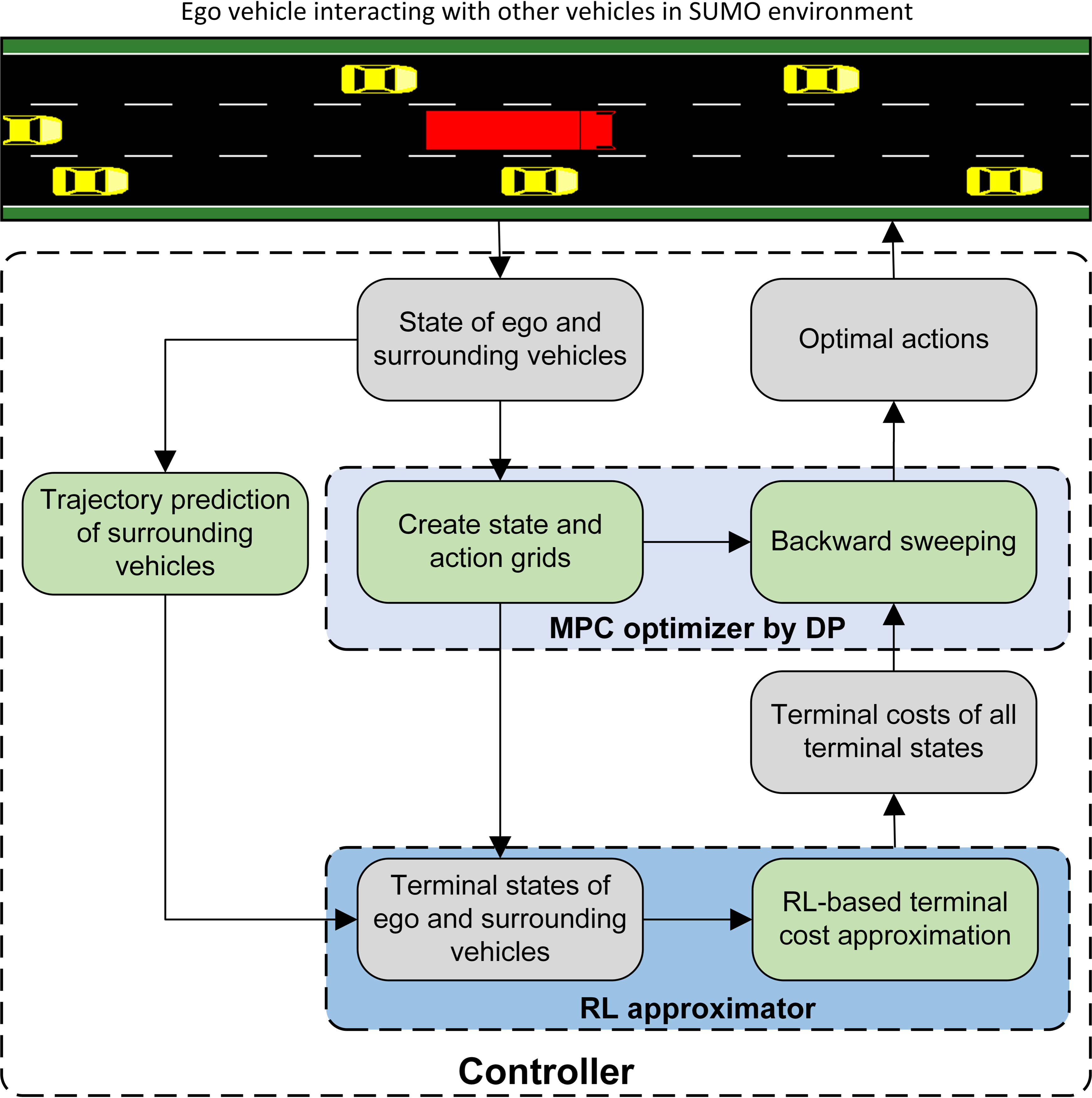}
    \caption{Architecture of controller}
    \label{Flowchart}
    \Description{The architecture of the proposed controller consists of three main operational tiers: the environment simulator, the state-prediction pipeline, and a two-stage hybrid decision-making block comprising an MPC optimizer and an RL approximator. At the top layer, the ego vehicle interacts with surrounding traffic within a SUMO (Simulation of Urban MObility) environment. The current simulator state yields data on the ego and surrounding vehicles, which branches into two parallel paths. The left pipeline focuses on future-horizon context, estimating the trajectory prediction of surrounding vehicles. Concurrently, the primary control block feeds the immediate traffic state into the MPC optimizer layer, where a discretized state-action grid is constructed. The predictive trajectory data and the current state-action grids converge at the bottom layer inside the RL approximator block. Here, the terminal states of both the ego and surrounding vehicles are evaluated. An RL-based network approximates the terminal cost for these configurations, outputting the terminal costs of all terminal states back to the MPC block. Finally, the MPC optimizer uses these values to execute a backward sweeping dynamic programming (DP) routine. This optimization loop completes by generating the sequence of optimal actions, passing the primary action back to update the SUMO environment.}
\end{figure}

Figure~\ref{Flowchart} illustrates the architecture of proposed hybrid method. At each step, the controller observes the current traffic state, constructs a discretized state-action grid, and evaluates the terminal cost, then runs the backward DP to recover the optimal action sequence. Only the first action is sent to simulator, and the loop repeats. For benchmarking, we also evaluate pure MPC, which sets $V_f = 0$; a rollout-based terminal cost, which simulates a base policy forward from the horizon boundary; and pure RL, which applies the learned policy directly without any MPC optimization.

\subsection{Trajectory Prediction}
\label{Trajectory Prediction}
 
Evaluating the safety terms of the cost in Section~\ref{Cost Function} over the prediction horizon requires an estimate of each surrounding vehicle's future position, speed, and lane at every horizon step $i=1,\ldots,N$, since their true future behavior is not known to the controller. To model typical highway driving behavior, surrounding vehicles are extrapolated forward from their currently observed states under a constant-acceleration assumption, holding each vehicle's lane fixed for the duration of the relatively short horizon. Among the prediction strategies evaluated, this approach achieves the lowest mean position error at all tested horizon lengths, attributed to the short timescales involved where the current acceleration is the dominant predictor of near-term position. 

However, this prediction is affected by the uncertainty in initial measurements. Each surrounding vehicle's initial position, speed, and acceleration are observed with additive Gaussian noise:

\begin{subequations}
\begin{align}
    &\tilde{s}_{j,0} = s_{j,0} + \nu_s, \quad \nu_s \sim \mathcal{N}(0,\sigma_s^2)\\
    &\tilde{v}_{j,0} = v_{j,0} + \nu_v, \quad \nu_v \sim \mathcal{N}(0,\sigma_v^2)\\
    &\tilde{a}_{j,0} = a_{j,0} + \nu_a, \quad \nu_a \sim \mathcal{N}(0,\sigma_a^2)
\end{align}
\end{subequations}
with $\nu_s, \nu_v, \nu_a$ independent. Propagating these through the constant-acceleration model yields a Gaussian distribution over the predicted position at horizon step $i$, with variance:

\begin{equation}
\label{eq:sigma_growth}
    \sigma_{s,i}^2 = \sigma_s^2 + (i\,\Delta t)^2\,\sigma_v^2
    + \left(\tfrac{i(i-1)}{2}(\Delta t)^2\right)^2\sigma_a^2
\end{equation}
so positional uncertainty grows with the horizon step, quadratically in the speed noise and quadratically in the acceleration noise.
 
A neighbor vehicle that is actively executing a lane change at the time of prediction is handled separately, since its true lane occupancy is ambiguous while it is physically transitioning between lanes. Rather than committing it to either lane, such a vehicle is represented by two copies in the predicted trajectory set, the original vehicle on its origin lane and a duplicate ghost vehicle, with identical position, speed, and acceleration, placed on its target lane. This ensures the safety and collision terms of $\ell$ account for the neighbor vehicle's presence in both lanes during the transition, preventing the controller from planning a maneuver into a gap that is already being closed by an in-progress lane change. More details of the tested prediction models are provided in Appendix~\ref{Trajectory Prediction Methods}. 

\subsection{Cost Function under Uncertainty}
\label{Cost Function}

The stage cost $\ell(x^e_i,u^e_i\mid x^T_i)$ combines seven terms into a single weighted sum:

\begin{equation}
    \ell(x^e_i, u^e_i \mid x^T_i) = \ell_\tx{speed} + \ell_\tx{accel} + \ell_\tx{LC}
    + \ell_\tx{CD} + \ell_\tx{safety} + \ell_\tx{LL} + \ell_\tx{col}
\end{equation}

Of these, only the safety and collision terms depend on the predicted states of surrounding vehicles, and therefore only these are affected by the uncertainty. The remaining terms are functions of the ego state alone and are unaffected.

The speed and acceleration terms shape the nominal driving behavior:

\begin{equation}
    \ell_\tx{speed} = w_v(v^e - v_\tx{ref})^2, \qquad
    \ell_\tx{accel} = w_a (a^e)^2
\end{equation}
where $v_\tx{ref}$ is the reference speed and $w_v, w_a > 0$ are tunable weights penalizing speed deviation and control effort.

A lane change is penalized once per maneuver, at the step it is initiated, using the counter $c^e$ that governs the transition lock:

\begin{equation}
    \ell_\tx{LC} =
    \begin{cases}
        P_\tx{LC} & \text{if } c^e \leq C_\tx{start}
                    \text{ and } \Delta l^e \neq 0\\
        0 & \text{otherwise}
    \end{cases}
\end{equation}
where $P_\tx{LC} > 0$ is a fixed penalty and $C_\tx{start}$ is the counter threshold below which a new lane change is permitted. A separate cooldown penalty $\ell_\tx{CD}$, with a larger fixed value $P_\tx{CD} > P_\tx{LC}$, discourages a second maneuver shortly after the first without making necessary back-to-back changes prohibitively costly. A progressive left-lane penalty $\ell_\tx{LL} = w_\tx{LL}\cdot l^e$, zero in the rightmost lane,
encourages the truck to return to the slow lane once an overtaking maneuver is no longer beneficial.

The safety and collision terms depend on the bumper-to-bumper gaps $g_\tx{front}$ and $g_\tx{back}$ to the relevant neighbors, which are computed from the predicted surrounding vehicles' states. Safety and collision avoidance requirement of Eq.~\eqref{eq:col_avoid} is imposed on both front and back gaps. For front gap:

\begin{equation}
    \mathbb{P}\bigl(\tilde{g}_{\tx{front},i} \geq d_\tx{safe}\bigr)
    \geq 1 - \epsilon
\end{equation}
where $d_\tx{safe} = D_0 + T_\tx{head}\,v^e$ is the speed-dependent safe distance, $D_0$ a minimum gap, and $T_\tx{head}$ a time headway. Since $\tilde{g}_{\tx{front},i}$ is Gaussian with standard deviation $\sigma_{s,i}$, this admits the deterministic equivalent:

\begin{equation}
\label{eq:d_safe_robust}
    d_\tx{safe}^\tx{robust}(i) = d_\tx{safe}
    + \Phi^{-1}(1-\epsilon)\cdot\sigma_{s,i}
\end{equation}
where $\Phi^{-1}$ is the inverse Gaussian CDF. Because $\sigma_{s,i}$ grows with $i$ by Eq.~\eqref{eq:sigma_growth}, the controller becomes progressively more conservative at later horizon steps. Similar logic applies to back gap.

Safety is enforced as a quadratic penalty when a gap falls below $d_\tx{safe}^\tx{robust}$, applied to the front vehicle at all times and to the rear vehicle on the target lane whenever a lane change is commanded:

\begin{equation}
\label{eq:safety_cost}
\begin{split}
    \ell_\tx{safety} =\; & w_\tx{safe}\Bigl(
      \max\bigl(0,\, d_\tx{safe}^\tx{robust}(i) - g_\tx{front}\bigr)^2 \\
    &+ \delta_\tx{lc}\,
      \max\bigl(0,\, d_\tx{safe}^\tx{robust}(i) - g_\tx{back}\bigr)^2
    \Bigr)
\end{split}
\end{equation}
where $w_\tx{safe} > 0$ and $\delta_\tx{lc} = 1$ if $\Delta l^e \neq 0$.

A large fixed penalty $\ell_\tx{col} = P_\tx{col}$ is charged whenever the front gap becomes non-positive, or whenever a commanded lane change would place the ego closer to a faster rear vehicle on the target lane than that vehicle could brake to match within its maximum deceleration. This braking-distance check is more conservative than a simple gap test, since a positive but rapidly closing rear gap can still result in a collision shortly after the maneuver completes. It is necessary because the DP assigns no back collision penalty to a vehicle already in the ego's lane, and penalizes rear risk on the target lane only during a commanded change. 

\subsection{Discounted Objective}
\label{Discounted Objective}

The discount factor $\gamma$ introduced in Eq.~\eqref{eq:optimization} departs from the undiscounted objective common in classical MPC, and is motivated by the structure of the uncertainty in this setting. As $i$ grows, measurement noise accumulates through the prediction model and the constant-acceleration assumption of Section~\ref{Trajectory Prediction} degrades, so costs evaluated near $i=N$ rest on weaker evidence than those near $i=1$. The factor $\gamma^i$ down-weights them accordingly, preventing the controller from over-committing to uncertain predictions.

The same factor $\gamma^N$ is applied to the terminal cost, so that $V_f$ is discounted consistently with the stage costs it summarizes. Using the same $\gamma$ in RL training additionally ensures that the learned $V_f$ and the accumulated stage costs are expressed on a common scale, which is required for the exactness argument of Section~\ref{Reinforcement Learning and the Bellman Equation}.

\subsection{RL-Based Terminal Cost Approximation}
\label{RL-Based Terminal Cost}

Following the MDP formulation of Section \ref{Reinforcement Learning and the Bellman Equation},  the agent observes a 14-dimensional state comprising the ego speed and lane index, bumper-to-bumper gaps and relative velocities to the nearest neighbor vehicle in the current lane and each adjacent lane, and two indicators encoding the phase of an in-progress lane change or cooldown. The action space is the same acceleration and lane-change pair used by the MPC. The reward is the negative of the MPC stage cost, $r_i=-r_\mathrm{scale}\,\ell(x^e_i,u^e_i \mid x^T_i)$, where $r_\mathrm{scale}$ is for that the learned value function and the MPC's own cost share a scale by construction. A Q-network is trained via standard DQN~\cite{mnih2015human} to approximate $Q^*$. DQN is used rather than an actor-critic method such as PPO because the quantity required at the horizon boundary is an estimate of the optimal cost-to-go. The identity $V^{*}(x)=\max_{u}Q^{*}(x,u)$ makes an off-policy $Q$-network an estimate of $V^{*}$ by construction, whereas an actor-critic critic estimates $V^{\pi}$ for the policy being learned, which would bias the DP toward that policy's behavior rather than the optimum of the MPC objective.
 
Given the reward definition above, the optimal Q-function satisfies:
\begin{equation}
    Q^*(x^e, u^e) \approx -r_\tx{scale}\;\mathbb{E}\!\left[\sum_{i=0}^{\infty} \gamma^i\,\ell(x^e_{k+i}, u^e_{k+i} \mid x^T_{k+i})\right]
\end{equation}
so $V^*(.)$ can be rescaled into the MPC's cost units as a terminal cost:
\begin{equation}
    \hat{V}_f(x^e_N\mid x^T_N) = -\frac{\max_{u}\, Q(x^e_N, u^e;\, \theta)}{r_\mathrm{scale}}
\end{equation}
where $\theta$ denotes the trained Q-network parameters. 
 
Since the DP state $x_N = [s^e_N, l^e_N, v^e_N, c^e_N]^\top$ and the RL observation $z_N \in \mathbb{R}^{14}$ use different representations, a mapping $z_N = \phi(x^e_N, \mathcal{T}_N)$ converts one into the other before the Q-network is queried. The mapping constructs $z_N$ from the predicted traffic state $\mathcal{T}_N = \{(s^T_{j,N}, v^T_{j,N}, l^T_{j,N})\}_j$ at horizon step $N$ by computing the relevant gaps, relative velocities, and lane-change indicators that form the 14-dimensional RL observation.

\subsection{Rollout-Based Terminal Cost Approximation}
\label{Rollout-Based Terminal Cost}

To provide a baseline for the RL-based approach, we additionally adapt the approach of \cite{bertsekas2012dp2} to our setting. Rather than tabulating a cost-to-go via DP over an extended horizon, the rollout terminal cost extends planning online. From each terminal state $x^e_N$, a base policy is simulated forward for $N_{r}$ further steps, and the resulting stage costs are accumulated into $\hat{V}_{f_r}(x^e_N \mid x^T_N)$. The base policy re-implements the Krauss car-following model~\cite{krauss1998microscopic} used internally by simulator for longitudinal propagation of surrounding vehicles, holding their lanes fixed. The rollout therefore does not invoke the simulator at runtime, instead using this lightweight re-implementation directly for $N_r$ steps. The ego vehicle additionally considers a lane change during the rollout when the gaps to the target lane's leader and follower are safe and the resulting speed is closer to the reference. 

Both the DP backward pass and the rollout apply the same discount factor $\gamma$, so a cost incurred at rollout step $r$ beyond the horizon carries the same weight it would under a longer single DP horizon, keeping the combined $N+N_{r}$-step estimate consistent across the MPC/rollout boundary.


\section{Results and Discussion}
\label{Results and Discussion}

\subsection{Experimental Setup}
\label{sec:experimental_setup}

The proposed framework is evaluated in a simulation environment developed using Simulation of Urban MObility (SUMO)~\cite{Lopez2018}, an open-source microscopic traffic simulator. The controller interacts with the environment through TraCI, SUMO's Traffic Control Interface, which provides surrounding vehicle states and accepts longitudinal and lateral commands for the ego vehicle at each step.

The scenario is a three-lane, one-way highway segment. The ego vehicle is a heavy-duty truck, and surrounding vehicles follow SUMO's Krauss car-following model~\cite{krauss1998microscopic} and LC2013 lane-change model~\cite{erdmann2015lane}, with desired speeds sampled uniformly within a fixed range, maintaining an approximately constant traffic density around the ego. Each episode runs for a fixed maximum duration or terminates on a collision. The dynamics and cost parameters are listed in Appendix~\ref{app:hparam_history}.

\subsection{RL Training}
\label{sec:rl_training}

\begin{figure}
    \centering
    \includegraphics[width=0.85\linewidth]{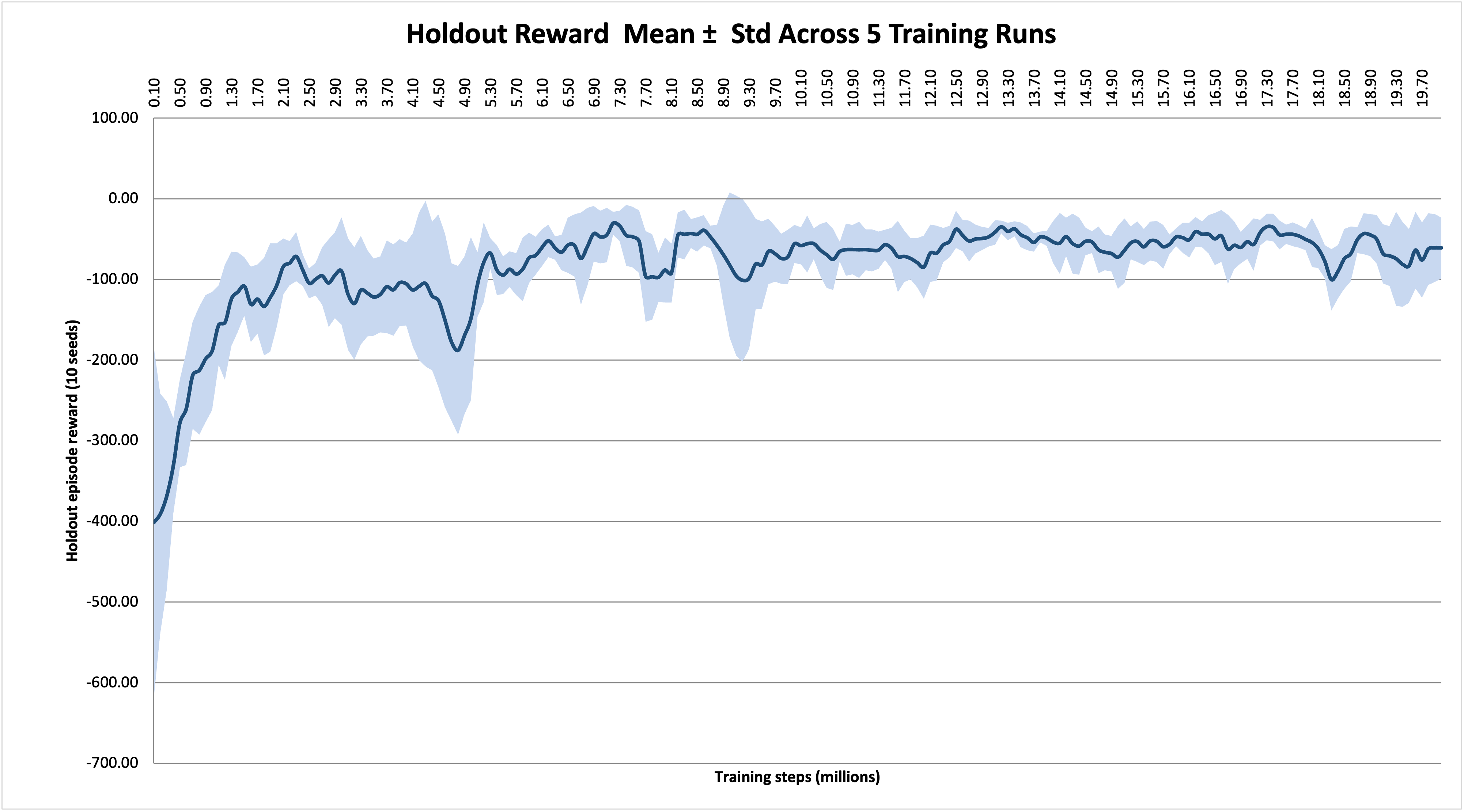}
    \caption{Holdout evaluation reward across five training runs, shown
    as mean $\pm$ one standard deviation.}
    \label{fig:multiseed_band}
    \Description{Line chart of holdout evaluation reward against
    training steps in millions, from 0 to 20 million. A solid dark line
    shows the mean across five independently trained DQN agents, and a
    shaded band around it spans one standard deviation either side. All
    runs rise steeply from a large negative reward during the first two
    to three million steps, then fluctuate around a common level for the
    remainder of training.}
\end{figure}

The DQN is trained using Stable-Baselines3~\cite{JMLR:v22:20-1364} in the deterministic simulation environment, under a range of hyperparameter settings summarized in Appendix~\ref{app:hparam_history}. Training with perfect state observations allows the agent to focus on learning long-term cost structure without the added variance of noisy measurements; the resulting value function is then used as a terminal cost within the stochastic MPC formulation, where measurement uncertainty is handled explicitly by the chance constraint. 

In every run, the model is checkpointed every 100k steps, and each checkpoint is evaluated on 10 holdout seeds with deterministic actions. Since the evaluation reward is dominated by rare, high-cost collision events, holdout performance varies between checkpoints, so a two-stage selection is used: the ten highest-scoring checkpoints are re-evaluated on 500 independent seeds, and the final model is chosen on mean reward and mean speed, the latter excluding policies that avoid collisions only by driving well below the reference speed. To assess reproducibility, the selected configuration was trained five times with this procedure. 

Figure~\ref{fig:multiseed_band} shows the holdout reward across the five runs. All runs improve rapidly within the first few million steps and fluctuate around a common level thereafter. Every run produced a collision-free policy at normal driving speed, with a mean reward of $-17.00 \pm 8.63$ and average speeds of $20.1$--$20.8\,\tx{m/s}$ across runs (Table~\ref{tab:multi_seed}, Appendix~\ref{app:hparam_history}). The model from the first run is used as the terminal cost throughout.

\subsection{Controller Comparison}
\label{sec:results}
The proposed hybrid MPC-RL controller and three baselines are evaluated on $100$ independent stochastic episodes. Deterministic results, obtained by suppressing measurement noise and setting $\epsilon \to 0.5$ which gives $\Phi^{-1}(1-\epsilon) = 0$ to recover the nominal predicted gaps, are reported in Appendix~\ref{app:det_results} as a baseline to isolate the contribution of uncertainty handling.

The proposed hybrid MPC-RL controller uses the learned value function to approximate the terminal cost as derived in Section~\ref{RL-Based Terminal Cost}, and is evaluated across horizons $N \in \{1,\ldots,5\}$. Pure MPC uses the same DP formulation with zero terminal cost ($V_f = 0$), isolating the effect of lookahead length alone. Rollout replaces the terminal cost with a rollout estimate from Section~\ref{Rollout-Based Terminal Cost}, evaluated at two rollout horizons $N_r \in \{10, 20\}$ to study how rollout length affects terminal cost quality. Pure RL applies the trained DQN policy directly with no MPC optimization, providing a reference for what the learned value function achieves without constraint handling or receding-horizon replanning. Deterministic variants of pure MPC, rollout ($N_r = 10$ and $20$), and hybrid MPC-RL apply noise-free conditions with full knowledge of the states of surrounding vehicles.

Table~\ref{tab:merged_stoc_results_detailed} reports collision rate, cost per step, cost standard deviation across all episodes ($\sigma_\tx{all}$), cost standard deviation across collision-free episodes only ($\sigma_0$), mean speed, and computation time per step, together with the full cost decomposition. Table~\ref{tab:det_vs_stoch} compares selected configurations against their deterministic counterparts.

\begin{table*}[t]
\centering
\caption{Stochastic controller performance evaluation and detailed cost decomposition.}
\label{tab:merged_stoc_results_detailed}
\resizebox{\textwidth}{!}{
\begin{tabular}{lcc cccccc ccccccc}
\toprule
& & & \multicolumn{6}{c}{\textbf{Performance Results}} & \multicolumn{7}{c}{\textbf{Detailed Cost Decomposition}} \\
\cmidrule(lr){4-9} \cmidrule(l){10-16}
Controller & $N$ & $N_r$ & Coll.\,\% & Cost/step & $\sigma_\tx{all}$ & $\sigma_0$ & Mean Speed & Time (ms) & $\ell_\tx{speed}$ & $\ell_\tx{accel}$ & $\ell_\tx{safety}$ & $\ell_\tx{col}$ & $\ell_\tx{LC}$ & $\ell_\tx{CD}$ & $\ell_\tx{LL}$\\
\midrule
Pure MPC      & 1 & -- & 86\% & 234{,}519 & 233{,}924 & 17{,}447 & 25.0 & 2.5   & 0    & 0    & 130{,}826 & 103{,}692 & 0.0286 & 0.0067 & 0.9703 \\
Pure MPC      & 2 & -- & 0\%  & 1{,}769   & 3{,}076   & 3{,}076  & 22.2 & 4.3   & 8.00 & 0.36 & 1{,}759   & 0         & 0.0149 & 0.0038 & 1.1031 \\
Pure MPC      & 3 & -- & 0\%  & 830       & 2{,}022   & 2{,}022  & 22.6 & 6.3   & 5.82 & 0.51 & 823       & 0         & 0.0147 & 0.0030 & 1.0109 \\
Pure MPC      & 4 & -- & 0\%  & 1{,}259   & 2{,}625   & 2{,}625  & 22.6 & 11.4  & 5.69 & 0.58 & 1{,}252   & 0         & 0.0136 & 0.0030 & 0.9717 \\
\textbf{Pure MPC} & \textbf{5} & \textbf{--} & \textbf{0\%} & \textbf{754} & \textbf{2{,}019} & \textbf{2{,}019} & \textbf{22.6} & \textbf{17.5} & \textbf{5.68} & \textbf{0.58} & \textbf{747} & \textbf{0} & \textbf{0.0133} & \textbf{0.0030} & \textbf{0.9629} \\
\midrule
Rollout  & 1 & 10 & 39\% & 65{,}159  & 74{,}003 & 4{,}740  & 22.6 & 9.1   & 5.56 & 0.60 & 3{,}606 & 61{,}546 & 0.0364 & 0.0284 & 0.5256 \\
\textbf{Rollout} & \textbf{2} & \textbf{10} & \textbf{0\%} & \textbf{1{,}072} & \textbf{3{,}058} & \textbf{3{,}058} & \textbf{22.5} & \textbf{15.1} & \textbf{6.17} & \textbf{0.65} & \textbf{1{,}065} & \textbf{0} & \textbf{0.0198} & \textbf{0.0052} & \textbf{0.4532} \\
Rollout  & 3 & 10 & 1\%  & 3{,}261   & 10{,}832 & 6{,}023  & 22.6 & 25.6  & 5.86 & 0.62 & 1{,}915 & 1{,}338  & 0.0208 & 0.0048 & 0.4511 \\
Rollout  & 4 & 10 & 2\%  & 3{,}978   & 13{,}137 & 5{,}021  & 22.6 & 41.5  & 5.85 & 0.61 & 1{,}286 & 2{,}686  & 0.0207 & 0.0070 & 0.4553 \\
Rollout  & 5 & 10 & 1\%  & 2{,}499   & 9{,}352  & 2{,}314  & 22.6 & 55.8  & 5.68 & 0.60 & 1{,}154 & 1{,}338  & 0.0216 & 0.0070 & 0.4478 \\
\midrule
Rollout  & 1 & 20 & 43\% & 65{,}150  & 73{,}027 & 4{,}265  & 22.7 & 12.8  & 5.31 & 0.62 & 4{,}701 & 60{,}443 & 0.0442 & 0.0396 & 0.6592 \\
Rollout  & 2 & 20 & 3\%  & 5{,}851   & 15{,}892 & 3{,}204  & 22.5 & 23.1  & 6.05 & 0.63 & 1{,}793 & 4{,}050  & 0.0242 & 0.0124 & 0.4769 \\
Rollout  & 3 & 20 & 4\%  & 6{,}395   & 18{,}937 & 2{,}447  & 22.7 & 37.1  & 5.51 & 0.60 & 1{,}652 & 4{,}736  & 0.0234 & 0.0091 & 0.4495 \\
Rollout  & 4 & 20 & 4\%  & 5{,}129   & 18{,}845 & 3{,}021  & 22.6 & 59.3  & 5.65 & 0.60 & 1{,}739 & 3{,}384  & 0.0224 & 0.0085 & 0.4636 \\
\textbf{Rollout} & \textbf{5} & \textbf{20} & \textbf{2\%} & \textbf{2{,}390} & \textbf{16{,}475} & \textbf{1{,}666} & \textbf{22.7} & \textbf{85.8} & \textbf{5.53} & \textbf{0.58} & \textbf{1{,}035} & \textbf{1{,}348} & \textbf{0.0239} & \textbf{0.0097} & \textbf{0.4646} \\
\midrule
\textbf{Pure RL} & \textbf{--} & \textbf{--} & \textbf{0\%} & \textbf{551}  & \textbf{1{,}029 } & \textbf{1{,}029 } & \textbf{20.8}  & \textbf{0.3} & \textbf{18.0} & \textbf{2.4} & \textbf{529} & \textbf{0} & \textbf{0.017} & \textbf{0.02} & \textbf{2.17} \\
\midrule
Hybrid   & 1 & -- & 0\%  & 1{,}475   & 1{,}692  & 1{,}692  & 21.1 & 29.1  & 15.25 & 2.55 & 1{,}454.34 & 0 & 0.0229 & 0.0184 & 3.0524 \\
Hybrid   & 2 & -- & 0\%  & 585       & 1{,}380  & 1{,}380  & 21.1 & 52.8  & 15.53 & 1.97 & 563.73   & 0 & 0.0205 & 0.0126 & 3.4646 \\
\textbf{Hybrid} & \textbf{3} & \textbf{--} & \textbf{0\%} & \textbf{329} & \textbf{1{,}488} & \textbf{1{,}488} & \textbf{21.1} & \textbf{91.6} & \textbf{14.90} & \textbf{1.80} & \textbf{309.01} & \textbf{0} & \textbf{0.0194} & \textbf{0.0112} & \textbf{3.4633} \\
Hybrid   & 4 & -- & 0\%  & 979       & 2{,}947  & 2{,}947  & 21.2 & 161.4 & 14.70 & 1.73 & 958.83   & 0 & 0.0218 & 0.0146 & 3.3872 \\
Hybrid   & 5 & -- & 1\%  & 1{,}872   & 8{,}333  & 2{,}173  & 21.2 & 176.0 & 14.85 & 1.67 & 1{,}184.33 & 668.11 & 0.0213 & 0.0114 & 3.2741 \\
\bottomrule
\end{tabular}
}
\end{table*}

\subsection{Discussion}
\label{sec:discussion}
 
\subsubsection{Effect of the terminal cost}
 
At a one-step horizon, pure MPC produces collisions in $86\%$ of the episodes. With a single optimization step, the controller minimizes only the immediate stage cost with no visibility of future states. When the ego truck approaches a slow leader, the speed-tracking term encourages maintaining $v_\tx{ref}$, but because the front safety cost grows quadratically with gap shortfall, the DP evaluates deceleration as expensive once the gap is already small. The only apparently cheaper option is a lane change, but the target lane may have its own slow leader, a situation the one-step DP cannot see. The cost decomposition confirms this. At $N=1$ the safety cost dominates at over $130{,}000$ and the collision cost reaches over $100{,}000$, while the speed cost is zero because the controller never deviates from $v_\tx{ref}$ before colliding. Adding one step of lookahead drops the collision rate to zero, since even $0.4\,\tx{s}$ of preview is sufficient for the DP to evaluate the consequence of not decelerating.
 
The hybrid controller at $N=1$ eliminates this failure mode entirely, achieving zero collisions. The learned terminal cost provides a long-term value signal that recognizes proximity to a slow leader as a high-cost state even when the immediate stage cost appears moderate, compensating for the absence of explicit future trajectory optimization. The hybrid at $N=3$ achieves the best overall result ($329$ cost, $0\%$ collisions), outperforming pure MPC at $N=5$ ($754$ cost) while using a substantially shorter horizon. The cost decomposition shows that the safety cost at hybrid $N=3$ ($309$) is $59\%$ lower than pure MPC $N=5$ ($747$), meaning the hybrid maintains larger safety margins on average, not just fewer collisions. This confirms the central claim of this work, namely that a learned terminal cost approximation achieves better driving quality at shorter, cheaper horizons than a zero terminal cost at longer ones. Acceleration and cooldown costs are negligible across all controllers and do not drive any of the observed differences.
 
The horizon also interacts with how effectively the controller handles prediction uncertainty. At $N=2$, the wider safety distances derived from the growing prediction uncertainty add conservatism that increases cost relative to shorter-horizon pure MPC, because the RL terminal value does not internally account for the tightened safety buffers it was not exposed to during training. At $N=3$, the additional headroom from prediction-uncertainty-aware safety distances corrects a failure mode that occurs at this horizon without significantly penalizing the already-conservative terminal value signal, yielding the best overall result. This suggests that a moderate horizon and prediction uncertainty-aware safety distances are mutually reinforcing, where the horizon provides enough look-ahead for the safety distances to be absorbed without sacrificing efficiency, while the safety distances correct edge cases the terminal cost alone
misses.
 
\subsubsection{Rollout horizon length and base policy quality}
 
The rollout controller with $N_r=10$ partially improves short-horizon safety. The collision rate at $N=1$ drops from $86\%$ to $39\%$, confirming that the rollout estimate introduces genuine long-term awareness. However, it does not eliminate collisions at any horizon, and the total cost remains up to $44$ times higher than the hybrid at the same $N$. The dominant cost terms are safety and collision, indicating that the rollout estimate is systematically overoptimistic. The base policy simulates a future in which the ego keeps a safe gap, but the actual closed-loop trajectory plays out differently because surrounding traffic evolves in ways the Krauss-based, no-lane-change rollout does not capture. The uncertainty margin partially compensate for this at $N=2$ by suppressing overoptimistic lane-change trajectories, which is why $N_r=10$ at $N=2$ achieves zero collisions.
 
Increasing the rollout horizon to $N_r=20$ consistently degrades performance. Collision rates are several times higher than for $N_r=10$ at every horizon beyond $N=1$, and cost per step is substantially larger. The discount factor reduces the weight of distant rollout steps, but the accumulated prediction mismatch over $20$ steps outweighs this correction. The DP is guided toward trajectories that appear favorable under a long rollout simulation
but prove unsafe in practice. This demonstrates that the quality of a rollout terminal cost is bounded by the accuracy of the base policy, not by the length of the simulation, and that extending the rollout horizon is counterproductive when the base policy is imperfect.
 
\subsubsection{Pure RL as a direct policy}
 
Pure RL achieves zero collisions and a total cost of $551$, competitive with pure MPC at $N=3$. The cost decomposition reveals a qualitatively different structure from the hybrid. The speed cost ($18.0$) is more than three times higher than for pure MPC or hybrid ($\approx5$ and $\approx15$ respectively), and the acceleration cost ($2.4$) is the highest of any safe controller, reflecting a jerky longitudinal profile compared to the DP-optimized trajectories. These observations are consistent with a known tendency of value-based policies. The sparse, large collision penalty dominates training, producing a policy that avoids collisions conservatively through large gaps and low speed rather than by optimizing the full multi-objective cost. This conservatism makes pure RL a poor stand-alone controller for a truck that must maintain traffic flow, but it is precisely what makes the learned value function effective as a terminal cost. It provides a cautious, risk-averse signal that discourages the hybrid DP from ending the horizon in states with insufficient safety margins.

\balance
 
\subsubsection{Comparison with deterministic baseline}

\begin{table}
\centering
\footnotesize
\caption{Deterministic vs.\ stochastic comparison for representative configurations.}
\label{tab:det_vs_stoch}
\begin{tabular}{lccccc}
\toprule
Controller & N & Det.\ Coll. & S.\ Coll. & Det.\ Cost & S.\ Cost \\
\midrule
Pure MPC          & 4 & 1\% & 0\%  & 1{,}590 & 1{,}259 \\
Pure MPC          & 5 & 0\% & 0\%  & 448     & 754     \\
Rollout $N_r\!=\!10$ & 2 & 1\% & 0\% & 2{,}078 & 1{,}072 \\
Rollout $N_r\!=\!10$ & 5 & 3\% & 1\% & 5{,}072 & 2{,}499 \\
Hybrid            & 2 & 0\% & 0\%  & 257     & 585     \\
Hybrid            & 3 & 1\% & 0\%  & 1{,}015 & 329     \\
\bottomrule
\end{tabular}
\end{table}
 
Table~\ref{tab:det_vs_stoch} compares selected configurations against their deterministic counterparts. The effect is not uniform. For configurations that already produce collisions in the deterministic setting, accounting for prediction uncertainty consistently reduces or eliminates collisions while also reducing cost, because the collision has big penalty and the wider safety distances suppress the overoptimistic trajectories that caused the deterministic failures. For configurations that are already collision-free deterministically, accounting for uncertainty sometimes increases cost due to the added conservatism of the wider safety distances. For example, the cost for hybrid controller at $N=2$ increases by $128\%$ ($257 \to 585$). Three configurations improve on both axes, becoming collision-free while also reducing cost: pure MPC at $N=4$, rollout $N_r{=}10$ at $N=2$, and the hybrid at $N=3$. Of these, the hybrid at $N=3$ shows by far the largest improvement ($1{,}015 \to 329$), making it the strongest configuration overall. This pattern confirms that the uncertainty-aware formulation is more conservative where predictions are uncertain and the deterministic controller would otherwise take risks.
 
\subsubsection{Computational cost}
 
All controllers remain within the $200\,\tx{ms}$ simulation-time budget, with the hybrid at $N=5$ the most expensive at $176.0\,\tx{ms}$. The hybrid controller's cost grows faster with horizon than pure MPC because each reachable terminal state requires a forward pass through the Q-network, which has two hidden layers of 256 units, in addition to the DP state-space expansion. The best-performing configuration, hybrid $N=3$, uses $91.6\,\tx{ms}$ per
step, under half the budget, so the recommended operating point carries a substantial timing margin. Batching terminal-state network evaluations or reducing the network size would lower this further and provide headroom for longer horizons on more constrained hardware. It should be noted that the reported times are influenced by the workload of the shared computational cluster at the time of evaluation.

\section{Conclusion}
\label{Conclusion}
 
This paper presented a hybrid MPC-RL framework for autonomous highway driving of a heavy-duty truck, in which a chance-constrained finite-horizon MPC solved over a discretized state-action space uses a DQN-learned approximation of the true cost-to-go as its terminal cost. The framework was evaluated against pure MPC, rollout-based, and pure RL baselines in a traffic simulator with reactive surrounding vehicles.

The results show that a learned terminal cost resolves the short horizon failure of pure MPC, achieves better driving quality at shorter horizons than a zero terminal cost at longer ones, and remains within real-time computational budget. Extending the rollout horizon degrades rather than improves performance, showing that rollout quality is bounded by base policy accuracy rather than simulation length.

Limitations include a straight homogeneous road, a kinematic ego model with no continuous lateral control, and a trajectory prediction model that assumes constant acceleration and fixed lanes for surrounding vehicles, which becomes less accurate at longer horizons and misses reactive lane-change behavior. Future work can address these by extending to richer road geometries, incorporating more accurate and reactive prediction models, and batching network evaluations to reduce computational cost at longer horizons.

\begin{acks}
This work was partially supported by the Wallenberg AI, Autonomous Systems
and Software Program (WASP) funded by the Knut and Alice Wallenberg Foundation.
\end{acks}


\bibliographystyle{ACM-Reference-Format} 
\bibliography{sample}

\newpage 
\appendix

\section{Deterministic Controller Results}
\label{app:det_results}
The deterministic evaluation of the controllers is obtained by suppressing measurement noise and setting $\epsilon \to 0.5$ so that $\Phi^{-1}(1-\epsilon) = 0$ and the tightened safe distance of Eq.~\eqref{eq:d_safe_robust} reduces to the nominal $d_\tx{safe}$. The controller then acts on the nominal predicted gaps with full knowledge of the surrounding vehicle states. All other settings, including the 100 evaluation episodes and the checkpoint used for the terminal cost, are identical to the stochastic evaluation of Section~\ref{sec:results}, so the two differ only in the treatment of uncertainty. Table~\ref{tab:det_results} reports the aggregate results and Table~\ref{tab:decomp} the corresponding cost decomposition. The cooldown penalties are negligible throughout, therefore are omitted from Table~\ref{tab:decomp}.

\begin{table}[h]
\centering
\footnotesize
\caption{Deterministic controller results.}
\label{tab:det_results}
\begin{tabular}{lcccccc}
\toprule
Controller & N & $N_r$ & Coll.\,\% & Cost/step & Mean Speed & Time/step (ms) \\
\midrule
Pure MPC        & 1 & -- & 88\% & 241{,}871 & 25.0  & 2.5  \\
Pure MPC        & 2 & -- & 0\%  & 1{,}084   & 22.1  & 4.3  \\
Pure MPC        & 3 & -- & 0\%  & 647       & 22.6  & 6.3  \\
Pure MPC        & 4 & -- & 1\%  & 1{,}590   & 22.6  & 11.4 \\
Pure MPC        & 5 & -- & 0\%  & 448       & 22.7  & 17.5 \\
\midrule
Rollout         & 1 & 10 & 34\% & 53{,}562 & 22.5 & 7.8  \\
Rollout         & 2 & 10 & 1\%  & 2{,}078  & 22.6 & 14.4 \\
Rollout         & 3 & 10 & 2\%  & 4{,}079  & 22.6 & 21.8 \\
Rollout         & 4 & 10 & 2\%  & 5{,}203  & 22.6 & 37.4 \\
Rollout         & 5 & 10 & 3\%  & 5{,}072  & 22.6 & 75.1 \\
\midrule
Rollout         & 1 & 20 & 32\% & 49{,}333 & 22.8 & 11.4  \\
Rollout         & 2 & 20 & 13\% & 12{,}735 & 22.6 & 23.1  \\
Rollout         & 3 & 20 & 10\% & 8{,}946  & 22.7 & 33.1  \\
Rollout         & 4 & 20 & 15\% & 12{,}363 & 22.6 & 59.8  \\
Rollout         & 5 & 20 & 7\%  & 8{,}942  & 22.8 & 122.0 \\
\midrule
Hybrid      & 1 & --  & 0\%  & 883      & 21.1  & 17.1  \\
\textbf{Hybrid}      & \textbf{2} & \textbf{--} & \textbf{0\%} & \textbf{257} & \textbf{21.1} & \textbf{49.8}  \\
Hybrid      & 3 & -- & 1\%  & 1{,}015  & 21.2  & 113.7 \\
Hybrid      & 4 & -- & 0\%  & 300      & 21.2  & 171.5 \\
Hybrid      & 5 & -- & 2\%  & 2{,}178  & 21.1  & 242.3 \\
\bottomrule
\end{tabular}
\end{table}

\begin{table}[h]
\centering
\footnotesize
\caption{Cost decomposition for each controller type (deterministic).}
\label{tab:decomp}
\begin{tabular}{lcccccccc}
\toprule
Controller & N & $N_r$ & $\ell_\mathrm{speed}$ & $\ell_\tx{accel}$ & $\ell_\tx{safety}$ & $\ell_\tx{col}$ & $\ell_\tx{LC}$ & $\ell_\tx{LL}$\\
\midrule
Pure MPC & 1 & -- & 0.0  & 0.0  & 131{,}885  & 109{,}984 & 0.027 & 1.00 \\
Pure MPC & 2 & -- & 8.2  & 0.1  & 1{,}075    & 0         & 0.013 & 0.98 \\
Pure MPC & 3 & -- & 5.8  & 0.2  & 640        & 0         & 0.013 & 1.00 \\
Pure MPC & 4 & -- & 5.8  & 0.2  & 916        & 667       & 0.013 & 0.96 \\
\textbf{Pure MPC} & \textbf{5} & \textbf{--} & \textbf{5.3} & \textbf{0.2} & \textbf{442} & \textbf{0} & \textbf{0.013} & \textbf{1.01} \\
\midrule
Rollout & 1 & 10 & 6.4  & 0.2  & 2{,}500    & 51{,}052  & 0.023 & 0.49 \\
\textbf{Rollout} & \textbf{2} & \textbf{10} & \textbf{6.0} & \textbf{0.2} & \textbf{1{,}403} & \textbf{667} & \textbf{0.018} & \textbf{0.36} \\
Rollout & 3 & 10 & 5.7  & 0.2  & 2{,}069    & 1{,}335   & 0.019 & 0.41 \\
Rollout & 4 & 10 & 5.8  & 0.2  & 3{,}851    & 2{,}012   & 0.018 & 0.46 \\
Rollout & 5 & 10 & 5.9  & 0.3  & 3{,}033    & 2{,}030   & 0.019 & 0.46 \\
\midrule
Rollout & 1 & 20 & 5.0  & 0.2  & 3{,}436    & 45{,}892  & 0.030 & 0.52 \\
Rollout & 2 & 20 & 5.8  & 0.2  & 2{,}068    & 10{,}660  & 0.022 & 0.45 \\
Rollout & 3 & 20 & 5.3  & 0.2  & 1{,}309    & 7{,}632   & 0.022 & 0.42 \\
Rollout & 4 & 20 & 5.6  & 0.2  & 1{,}700    & 10{,}656  & 0.022 & 0.44 \\
\textbf{Rollout} & \textbf{5} & \textbf{20} & \textbf{5.0} & \textbf{0.2} & \textbf{3{,}441} & \textbf{5{,}496} & \textbf{0.021} & \textbf{0.49} \\
\midrule
Hybrid & 1 & -- & 15.4 & 1.4  & 863        & 0         & 0.024 & 3.13 \\
\textbf{Hybrid} & \textbf{2} & \textbf{--} & \textbf{15.3} & \textbf{1.3} & \textbf{237} & \textbf{0} & \textbf{0.017} & \textbf{3.61} \\
Hybrid & 3 & -- & 14.6 & 1.3  & 325        & 671       & 0.018 & 3.53 \\
Hybrid & 4 & -- & 14.9 & 1.3  & 280        & 0         & 0.016 & 3.46 \\
Hybrid & 5 & -- & 14.9 & 1.2  & 821        & 1{,}338   & 0.023 & 3.32 \\
\bottomrule
\end{tabular}
\end{table}

\section{Trajectory Prediction Methods} 
\label{Trajectory Prediction Methods}
Two strategies were considered for predicting the future states of surrounding vehicles over the MPC horizon.
 
\textbf{Reactive Multi-Agent Prediction.} The first strategy propagates surrounding vehicles forward using a re-implementation of SUMO's internal car-following logic. At each prediction step $i$, every neighbor vehicle adjusts its speed based on its gap to the vehicle immediately ahead of it, following the Krauss safe-speed model~\cite{krauss1998microscopic}. The ego vehicle is projected at constant speed and fixed lane to serve as a stable reference for those reactions, avoiding a circular dependency between the controller's own plan and the neighbors' predicted responses.
 
While this approach captures realistic following behavior in principle, it introduces a systematic error at the boundary of the sensor range. The lead vehicle in each lane, the frontmost vehicle visible to the ego, has no visible leader within the sensor range, so the car-following model treats it as having a clear road ahead and accelerates it toward its reference (maximum) speed. This artificial acceleration propagates backward through the platoon, producing growing position errors over longer horizons. Freezing the boundary leader at constant speed mitigates this to some extent, but compounding model-mismatch errors still accumulate at every step as small inaccuracies in each vehicle's response to the ego's projected state propagate forward through the chain.
 
\textbf{Constant-Acceleration Extrapolation.} The second strategy extrapolates each surrounding vehicle forward from its currently observed position $s_{j,0}$, speed $v_{j,0}$, and acceleration $a_{j,0}$, holding the lane assignment fixed:
\begin{align}
    s_{j,i} &= s_{j,0} + i\,v_{j,0}\,\Delta t + \tfrac{i(i-1)}{2}\,a_{j,0}\,(\Delta t)^2 \\
    v_{j,i} &= \mathrm{clip}\!\left(v_{j,0} + i\,a_{j,0}\,\Delta t, \; 0,\; v_{j,\max}\right)
\end{align}
This model requires no knowledge of the neighbor vehicles' car-following parameters and introduces no inter-vehicle dependencies, so boundary effects do not arise. 

\textbf{Comparison.} Both methods were evaluated by comparing predicted positions against ground-truth SUMO positions at each horizon step $i = 1,\ldots,5$, across $100$ seeds using the pure MPC controller. The mean position error (bias) and its standard deviation are reported in Table~\ref{tab:pred_comparison} for the longest horizon ($i=5)$, which has the most error.

\begin{table}[h]
\centering
\footnotesize
\caption{Mean ($\mu$) and standard deviation ($\sigma$) of position prediction error (m) for $i=5$.}
\label{tab:pred_comparison}
\begin{tabular}{lccccc}
\toprule
& \multicolumn{2}{c}{Reactive} & & \multicolumn{2}{c}{Const. acc.} \\
\cmidrule(lr){2-3} \cmidrule(lr){5-6}
Direction & $\mu$ & $\sigma$ & & $\mu$ & $\sigma$ \\
\midrule
Front       & 0.031 & 0.222 & & -0.002 & 0.089 \\
Front-left  & 0.079 & 0.295 & & 0     & 0.112 \\
Front-right & 0.059 & 0.262 & & 0.001 & 0.096 \\
Back-left   & 0.243 & 0.436 & & 0.005 & 0.133 \\
Back-right  & 0.167 & 0.393 & & 0     & 0.131 \\
\bottomrule
\end{tabular}
\end{table}

Constant-acceleration extrapolation achieves lower mean absolute error and standard deviation than the reactive model at all horizon steps. At $i=5$ ($1\,\text{s}$ ahead), the standard deviation remains below $0.15\,\text{m}$, confirming that the current acceleration is the dominant predictor of near-term position at the short timescales used by the DP solver. Constant-acceleration extrapolation is therefore adopted as the prediction model throughout this work.

\section{Rollout Base Policy}
\label{app:rollout}
 
The rollout base policy propagates the ego vehicle and surrounding vehicles forward for $N_r$ steps using lightweight analytical models, avoiding invocation of the SUMO simulator at runtime.
 
\textbf{Surrounding vehicles.} Each surrounding vehicle $j$ is propagated longitudinally using the Krauss safe-speed model~\cite{krauss1998microscopic}, holding its lane fixed throughout the rollout. The safe speed for vehicle $j$ following its leader is:
\begin{equation}
    v^j_\tx{safe} = -b\tau + \sqrt{(b\tau)^2 + (v^j_\tx{lead})^2 + 2b\,g^j}
\end{equation}
where $b$ is the maximum deceleration, $\tau$ is the reaction time, $v^j_\tx{lead}$ is the leader's speed, and $g^j$ is the gap to the leader. The vehicle then accelerates toward its desired speed subject to $v^j_\tx{safe}$.
 
\textbf{Ego vehicle.} For the ego vehicle, the safe speed is computed using the
speed-dependent safe distance:
\begin{equation}
    d_\tx{safe} = D_0 + T_\tx{head}\,v^e
\end{equation}
where $D_0$ is a minimum gap and $T_\tx{head}$ is a time headway, applied to the front vehicle at all times and to the rear vehicle on the target lane whenever a lane change is commanded. At each rollout step, the ego considers a lane change if three conditions are met simultaneously: the gap to the target lane's leader exceeds $d_\tx{safe}$; the gap to the target lane's follower exceeds $d_\tx{safe}$; and the achievable speed in the target lane is closer to $v_\tx{ref}$ than in the current lane. If conditions are met, the lane change is executed; otherwise the ego remains in its current lane. The ego state is then updated as:
\begin{align}
    v^e_{r+1} &= \mathrm{clip}(v^e_r + a^e_r\,\Delta t,\; v^e_\tx{min},\; v^e_\tx{max}) \\
    s^e_{r+1} &= s^e_r + v^e_{r+1}\,\Delta t
\end{align}
where $r$ indexes rollout steps and $a^e_r$ is computed to reach the target safe speed. The accumulated discounted stage cost over the rollout is
\begin{equation}
    \hat{V}_{f_r}(x^e_N) = \sum_{r=0}^{N_r - 1} \gamma^r\,\ell(x^e_{N+r},\, \pi_r(x^e_{N+r}))
\end{equation}
where $\pi_r$ is the rollout policy described above. Surrounding vehicle lane changes during the rollout horizon are not modeled, which makes the rollout optimistic in scenarios where a neighboring vehicle cuts in during the rollout extension.

\section{RL Model and Simulation Hyperparameter Tuning}
\label{app:hparam_history}

\begin{table}[h]
\footnotesize
\centering
\caption{Selected model of each training run for the final hyperparameter setting, evaluated on 500 seeds.}
\label{tab:multi_seed}
\begin{tabular}{lcccc}
\toprule
Run & Checkpoint & Collision \% & Mean reward & $v_{\text{avg}}$ (m/s) \\
\midrule
1 & 16.1M & 0 & $-31.46 \pm 42.44$ & 20.8 \\
2 & 12.6M & 0 & $-16.90 \pm 28.45$ & 20.1 \\
3 & 6.8M  & 0 & $-13.72 \pm 22.03$ & 20.5 \\
4 & 13.6M & 0 & $-8.60 \pm 20.39$  & 20.8 \\
5 & 17.5M & 0 & $-14.30 \pm 23.85$ & 20.6 \\
\midrule
Across runs & -- & 0 & $-17.00 \pm 8.63$ & $20.6$ \\
\bottomrule
\end{tabular}
\end{table}

\begin{table}[h]
\centering
\footnotesize
\setlength{\tabcolsep}{4pt}
\caption{MPC/DP solver, DQN network, and training parameters.
         Bold denotes selected value; parentheses list other
         values tested.}
\label{tab:params_algo}
\begin{tabular}{lc}
\toprule
\textbf{Parameter} & \textbf{Value} \\
\midrule
\textit{MPC / DP solver} & \\
~~Prediction horizon $N$                            & $1$--$5$ \\
~~Rollout horizon $N_r$                             & $10$,~$20$ \\
~~Discount factor $\gamma$                          & $0.98$ \\
~~Velocity grid step $V_\tx{step}$                  & $0.2\,\text{m/s}$ \\
~~Acceleration levels                               & $5$ \\
\midrule
\textit{DQN network} & \\
~~Hidden layers                                     & $[256,\,256]$ \\
~~Replay buffer size                                & $400{,}000$ \\
~~Batch size                                        & $64$ \\
~~Target network update                             & $10{,}000\,\text{steps}$ \\
\midrule
\textit{DQN training (tuned)} & \\
~~Discount factor $\gamma$      & $\mathbf{0.98}$ ($0.90$, $0.95$, $0.995$) \\
~~Speed weight $w_v$            & $\mathbf{1}$ ($3$, $5$, $10$) \\
~~Learning rate                 & $\mathbf{10^{-5}}$ ($10^{-4}$) \\
~~Sensor range $d_\tx{sensor}$  & $\mathbf{150\,\text{m}}$ ($100\,\text{m}$) \\
~~Expl.\ rate $\varepsilon_\tx{end}$ & $\mathbf{0.01}$ ($0.001$) \\
~~Checkpoint                    & $16.1\,\text{M}$ steps \\
\bottomrule
\end{tabular}
\end{table}

\begin{table}
\centering
\footnotesize
\setlength{\tabcolsep}{4pt}
\caption{Simulation, vehicle, and MPC cost parameters.}
\label{tab:params_sim}
\begin{tabular}{lc}
\toprule
\textbf{Parameter} & \textbf{Value} \\
\midrule
\textit{Simulation} & \\
~~Time step $\Delta t$                              & $0.2\,\text{s}$ \\
~~Episode length                                    & $300\,\text{s}$ \\
~~Lanes $N_\tx{lanes}$                              & $3$ \\
~~Traffic density                                   & $30\,\text{veh/km}$ \\
~~Surrounding speed range                           & $[20,\,30]\,\text{m/s}$ \\
~~Sensor range $d_\tx{sensor}$                      & $150\,\text{m}$ \\
\midrule
\textit{Ego vehicle} & \\
~~Speed limits $[v^e_\tx{min},\,v^e_\tx{max}]$     & $[0,\,33.33]\,\text{m/s}$ \\
~~Reference speed $v_\tx{ref}$                      & $25\,\text{m/s}$ \\
~~Acceleration $[a_\tx{min},\,a_\tx{max}]$          & $[-2.0,\,2.0]\,\text{m/s}^2$ \\
~~Vehicle length                                    & $12\,\text{m}$ \\
~~Lane-change duration $T_\tx{LC}$                  & $2.0\,\text{s}$ \\
\midrule
\textit{MPC cost function} & \\
~~Speed weight $w_v$                                & $1$ \\
~~Acceleration weight $w_a$                         & $1$ \\
~~Safety weight $w_\tx{safe}$                       & $1000$ \\
~~Left-lane weight $w_\tx{LL}$                      & $2$ \\
~~Min gap $D_0$                                     & $5\,\text{m}$ \\
~~Time headway $T_\tx{head}$                        & $2.0\,\text{s}$ \\
~~Lane-change penalty $P_\tx{LC}$                   & $10$ \\
~~Cooldown penalty $P_\tx{CD}$                      & $30$ \\
~~Collision penalty $P_\tx{col}$                    & $10^8$ \\
\bottomrule
\end{tabular}
\end{table}

Table~\ref{tab:multi_seed} lists the model selected from each of the five training runs of Section~\ref{sec:rl_training}, evaluated on 500 seeds. The configuration used in Section~\ref{sec:rl_training} was reached through the sequence of experiments summarized in Table~\ref{tab:params_algo}, and Tables~\ref{tab:params_sim} list the simulation and MPC cost parameters. Each setting was trained and screened on the 10 holdout seeds; only the final configuration was carried through the full 500-seed evaluation, so the outcomes below are qualitative and are not directly comparable to Table~\ref{tab:multi_seed}.

Training initially used a discount factor of $\gamma = 0.995$, which produced policies that avoided collisions by decelerating to a near standstill. Lowering the discount factor to $0.95$ and $0.90$ removed the degenerate behavior but left training unstable, with no checkpoint surviving the 500-seed evaluation; $\gamma = 0.98$ was retained. 

The learning rate was reduced from $10^{-4}$ to $10^{-5}$, which noticeably reduced the amplitude of the oscillation in the holdout curve without eliminating it. The sensor range was increased from $100$~m to $150$~m, and the final exploration rate was set to $\varepsilon_{\text{end}} = 0.01$; a lower value of $0.001$ produced comparable holdout results but a higher proportion of degenerate checkpoints.

The speed weight $w_v$ deserves particular attention. Values of $3$, $5$, and $10$ were tested, and while all trained stably, they degrade the hybrid controller rather than the RL policy itself. Because the reward is defined as the negative MPC stage cost (Section~\ref{RL-Based Terminal Cost}), any $w_v$ differing from the value used by the MPC rescales the learned $Q$-values relative to the stage costs accumulated over the horizon, so the terminal cost is no longer expressed in the same units as the objective it is inserted into. The speed weight in training must therefore match the MPC cost function, i.e.\ $w_v = 1$, independently of which value trains best in isolation.


\end{document}